\documentclass[10pt,twocolumn,letterpaper]{article}

\usepackage{cvpr}
\usepackage{times}
\usepackage{epsfig}
\usepackage{graphicx}
\usepackage{amsmath}
\usepackage{amssymb}

\usepackage[breaklinks=true,bookmarks=false]{hyperref}

\cvprfinalcopy 
\def\cvprPaperID{****} 

\begin{document}

\title{Audio Visual Scene-Aware Dialog (AVSD) Challenge at DSTC7
}

\author{
  Huda Alamri$^{*\dagger}$,
  Vincent Cartillier$^{*}$,
  Raphael Gontijo Lopes$^{*}$,
  Abhishek Das$^{*}$, 
  Jue Wang$^{\dagger}$,\\
  Irfan Essa$^{*}$,
  Dhruv Batra$^{*}$, 
  Devi Parikh$^{*}$, \\
  Anoop Cherian$^{\dagger}$,
  Tim K. Marks$^{\dagger}$,  
  Chiori Hori$^{\dagger}$\\[8pt]
$^{*}$School of Interactive Computing, Georgia Tech\\
$^{\dagger}$Mitsubishi Electric Research Laboratories (MERL), Cambridge, MA, USA \\
}

\maketitle
\begin{abstract}
  Scene-aware dialog systems will be able
  to have conversations with users about the objects and events around them. Progress on such systems can be made by integrating state-of-the-art technologies from multiple research areas including end-to-end dialog systems
  visual dialog,and video description. We introduce the Audio Visual Scene-Aware Dialog (AVSD) challenge and dataset. In this challenge, which is one track of the 7th Dialog System Technology Challenges (DSTC7) workshop\footnote{http://workshop.colips.org/dstc7/call.html}, the task is to build a system that generates responses in a dialog about an input video. 
\end{abstract}

\section{Introduction}
Spoken dialog technologies are becoming more common in real-world human-machine interfaces.
Recently, end-to-end training of neural networks has been shown to be a promising approach for training dialogue systems from  human-to-human dialogue corpora~\cite{DBLP:journals/corr/DasKGSYMPB16, lowe2015ubuntu}.
%
A variety of neural conversation models were tested at DSTC6~\cite{DBLP:journals/corr/HoriH17}. 
However, current dialog systems are unable to have a conversation about what is going on in the user's surroundings. The AVSD Challenge is motivated by the need for scene-aware dialog technology, so that machines can carry on a conversation with users about objects and events around them.

Entries to this challenge could capitalize on recent developments in related areas such as video description and Visual Dialog. 
Encoder-decoder networks developed for image captioning have recently been extended to the task of automatic {\em video description}~\cite{venugopalan2014translating}, which is the generation of natural language descriptions of videos (e.g., a sentence that summarizes an input video). To enhance video description performance, \cite{Hori_2017_ICCV} introduced an attention-based multimodal fusion approach that selectively attends to different input modalities such as audio and video features. 
Visual Dialog~\cite{DBLP:journals/corr/DasKGSYMPB16,visdial_rl,de2017guesswhat} 
extends visual question answering (VQA)~\cite{VQA} from simple single-turn question answering to multi-turn dialog, in which the utterances in each turn may reference information from previous turns of the dialog.

In the AVSD challenge, we further extend Visual Dialog by extending the subject of the interaction from unimodal static images to multimodal videos, where input features could come from multiple domains including image features, motion features, non-speech audio, and speech audio.

\section{Audio Visual Scene-Aware Dialog Challenge}
\subsection{Tasks}
In this challenge, the system must generate responses to a user input in the context of a given dialog. This context consists of a dialog history (previous utterances by both user and system) in addition to video and audio information that comprise the scene. The quality of a system's automatically generated sentences is evaluated using objective measures to determine whether or not the generated responses are natural and informative.

There are two tasks, each with two versions (a and b):
\begin{description} 
  \item[Task 1: Video and Text] (a) Use the video and text training data provided but no external data sources, other than publicly available pre-trained feature extraction models. (b) External data may also be used for training.
  \item[Task 2: Text Only] (a) Do not use the input videos for training or testing. Use only the text training data (dialogs and video descriptions) provided. (b) Any publicly available text data may also be used for training.
\end{description}
Challenge participants can select to submit entries in any or all of Task 1(a,b) and Task 2(a,b).  Training data and a baseline system will be released to all participants of DSTC7.

The quality of the automatically generated sentences will be evaluated with objective measures 
to measure the similarity between the generated sentences and ground truth sentences. 
We will use \texttt{nlg-eval}\footnote{https://github.com/Maluuba/nlg-eval} 
for objective evaluation of system outputs.

\subsection{Data collection}

We are collecting text-based human dialog data for videos from human action recognition datasets such as CHARADES\footnote{\url{http://allenai.org/plato/charades/}} and Kinetics\footnote{https://deepmind.com/research/open-source/open-source-datasets/kinetics/}.
We have already collected text-based dialog data about short videos 
from CHARADES~\cite{sigurdsson2016hollywood}, which contains untrimmed and multi-action videos, along with video descriptions.

The data collection paradigm for dialogs was similar to that described in~\cite{DBLP:journals/corr/DasKGSYMPB16}, in which for each image, two different Amazon Mechanical Turk (AMT) workers chatted via a text interface to yield a dialog. In~\cite{DBLP:journals/corr/DasKGSYMPB16},
each dialog consisted of a sequence of questions and answers about an image.
In our dataset, two AMT workers had a discussion about events in a video.
One of the workers played the role of an answerer who had already watched the video. 
The answerer answered questions asked by another AMT worker, the questioner.

The questioner was not shown the video but was only shown three static images: 
 the first, middle and last frames of the video.
Having seen static frames from the video, the questioner already has good information about image- and appearance-based information in the video. Thus, rather than focusing on scene information that is available in the static images, the dialog instead revolves around the events and other temporal features in the video, which is the content of interest for our AVSD dataset.
After 10 rounds of Q/A about the events that happened in the video, the questioner (who has not seen the video) is required to write a video description summarizing the events in the video.

In total, we have collected dialogs for 7043 videos from the CHARADES training set plus 
1465 videos from the validation set.
See Table~\ref{tab:data} for statistics.

\begin{table}[h]
\centering
\caption{\footnotesize{Audio Visual Scene-Aware Dialog Dataset on CHARADES. Since we did not have scripts for the test set,
we split the validation set into 732 and 733 videos and use them as our validation and test sets, respectively.}}
\label{tab:data}
\begin{tabular}{ll|ccc}
\hline
& 
& training 
& validation 
& test \\
\hline
\# of dialogs 
& 
& 7043
& 732  
& 733 \\
\# of turns   
& 
& 123,480
& 14,680
& 14,660
\\
\# of words   
& 
&  1,163,969
&  138,314
&  138,790
\\
\hline
\end{tabular}
\end{table}

\begin{figure}[thb]
    \centering
    \centerline{\includegraphics[width=9.0cm]{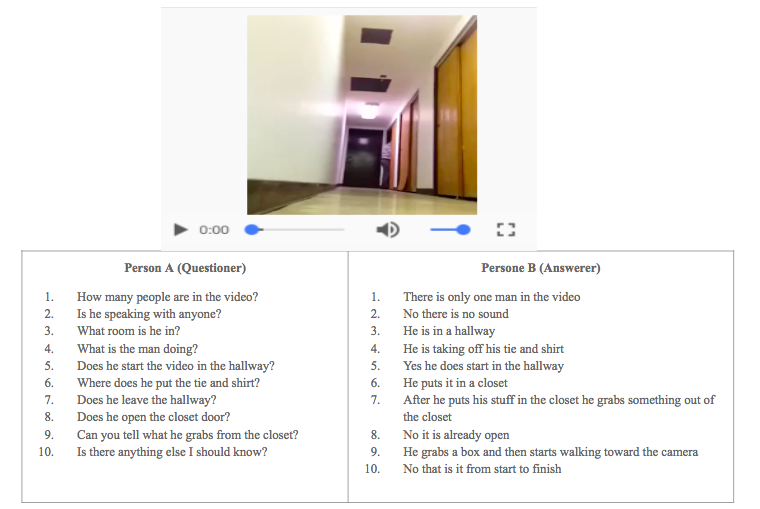}}
    \caption{A sample from our Audio Visual Scene-Aware Dialog (AVSD) dataset. The task of Scene-aware Dialog requires an agent to generate a meaningful response about a video in the context of the dialog.}
	\label{fig:avsd}
\end{figure}

\section{Summary}
We introduce a new challenge task and dataset---Audio Visual Scene-Aware Dialog (AVSD)---that form the basis of one track of the 7th Dialog System Technology Challenges (DSTC7) workshop. We collected human dialog data for videos from the CHARADES dataset and plan to collect more for videos from the Kinetics dataset. The information provided to participants will include a detailed description of the baseline system, instructions for submitting results for evaluation, and details of the evaluation scheme.

{\small
\bibliographystyle{ieee}
\bibliography{bibi}
}
\end{document}